\documentclass[11pt]{article}

\usepackage[final]{acl}

\usepackage{times}
\usepackage{latexsym}

\usepackage[T1]{fontenc}

\usepackage[utf8]{inputenc}

\usepackage{microtype}

\usepackage{inconsolata}

\usepackage{graphicx}
\usepackage{booktabs}
\usepackage{amsmath}
\usepackage{amssymb}

\usepackage{epsfig}
\usepackage{comment}
\usepackage{epigraph}
\usepackage{multirow}
\usepackage{hhline}
\usepackage{makecell}
\usepackage{bm}
\usepackage{subcaption}
\usepackage{soul}
\usepackage{array}
\usepackage{color,colortbl}
\usepackage{xcolor}

\makeatletter
\def\blfootnote{\gdef\@thefnmark{}\@footnotetext}
\makeatother

\definecolor{beaublue}{rgb}{0.74, 0.83, 0.9}

\title{TemporalVLM: Video LLMs for Temporal Reasoning in Long Videos}

\author{Fawad J. Fateh$^\dagger$~~~Umer Ahmed$^\dagger$~~~Hamza Khan~~~M. Zeeshan Zia~~~Quoc-Huy Tran\\
Retrocausal, Inc.\\
Redmond, WA\\
\url{www.retrocausal.ai}}

\begin{document}

\maketitle

\begin{abstract}
We introduce TemporalVLM, a video large language model (video LLM) for temporal reasoning and fine-grained understanding in long videos. Our approach includes a visual encoder for mapping a long-term video into features which are time-aware and contain both local and global cues. It first divides an input video into short-term clips, which are jointly encoded with timestamps and fused across overlapping temporal windows into time-sensitive local features. Next, the local features are passed through a bidirectional long short-term memory (BiLSTM) module for global feature aggregation. Moreover, to facilitate the evaluation of TemporalVLM, we present a large-scale long video dataset of industry assembly processes, namely IndustryASM, consisting of videos recorded on factory floors with actions and timestamps annotated by industrial engineers for time and motion studies and temporal action segmentation evaluation. Finally, extensive experiments show that TemporalVLM outperforms previous methods across temporal reasoning and fine-grained understanding tasks, i.e., dense video captioning, temporal video grounding, video highlight detection, and temporal action segmentation. To our best knowledge, our work is the first to incorporate LSTMs into video LLMs.
\end{abstract}

\section{Introduction}
\label{sec:introduction}
{\blfootnote{$^{\dagger}$ indicates joint first author.\\ \{fawad,umer,hamza,zeeshan,huy\}@retrocausal.ai.}}

\begin{figure*}[t]
	\centering
		\includegraphics[width=0.9\linewidth, trim = 0mm 75mm 0mm 0mm, clip]{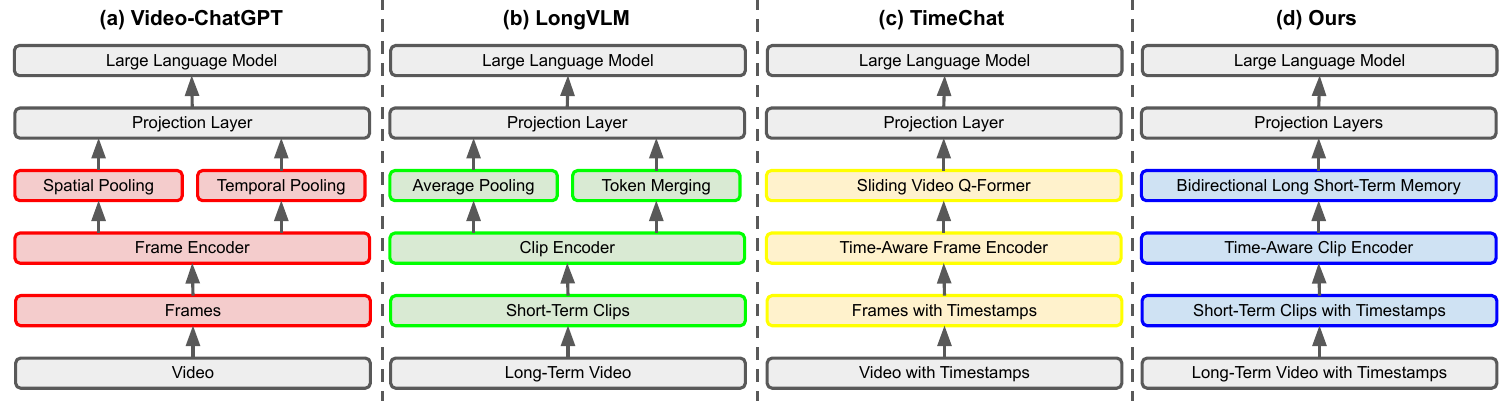}
    \captionof{figure}{Video LLMs are often not time-sensitive (a, b), consider an input video as a single clip (a, c), and apply pooling (a, b) or query aggregation (c) for aggregating global features. Our model (d) includes a time-aware clip encoder for extracting time-aware fine-grained cues and a BiLSTM for capturing long-range temporal dependencies.}
	\label{fig:teaser}
\end{figure*}

Video temporal reasoning represents the process of reasoning about time and its passage in videos, with a focus on how events or actions happen and relate to each other in terms of time. Several video understanding applications require the ability of temporal reasoning, e.g., dense video captioning~\cite{wang2021end,zhu2022end,yang2023vid2seq} and temporal action segmentation~\cite{li2020ms,yi2021asformer,hyder2024action,lu2024fact}. Considerable efforts~\cite{yang2023vid2seq,luo2023towards,moon2023query,zhong2023learning,lu2024fact,lin2024vedit} have been invested in developing models for solving individual tasks. These models often have different architectures. Thus, it is favorable to design a unified model for handling various tasks.

The past few years have witnessed the impressive comprehension and generation capability of LLMs~\cite{achiam2023gpt,vicuna2023,taori2023alpaca,openai_chatgpt,touvron2023llama,touvron2023llama2}, which have emerged as a universal agent for performing various tasks. Video LLMs~\cite{li2023videochat,liu2023video,luo2023valley,maaz2023video,zhang2023video,jin2024chat,song2024moviechat} which incorporate video encoders with LLMs have been introduced. These methods often represent a video by a fixed number of tokens, yielding reduced performance with long videos, and encode frames and timestamps separately and hence struggle with temporal reasoning tasks. Please refer to the inferior results of these methods in Tab.~\ref{table:zeroshot}. Recently, a few works~\cite{qian2024momentor,huang2024vtimellm,huang2024lita,ren2024timechat} have developed video LLMs with temporal reasoning abilities, e.g., TimeChat~\cite{ren2024timechat} proposes to vary the number of tokens based on the video length and jointly encode frames and timestamps. The above methods usually treat the entire video as a single clip~\cite{qian2024momentor,huang2024vtimellm,huang2024lita,ren2024timechat} and aggregate tokens via pooling operation~\cite{huang2024lita} and query aggregation~\cite{ren2024timechat}, struggling to capture fine-grained details in long videos.

We propose TemporalVLM, a video LLM for temporal reasoning and fine-grained understanding in long videos. Fig.~\ref{fig:teaser} shows the comparisons with prior works. Our model includes two architectural contributions. Firstly, we divide a long-term input video into multiple short-term clips and introduce a time-sensitive clip encoder for extracting fused time-aware local features from each clip. Secondly, we adopt a BiLSTM module which takes all the local features as inputs and computes global features from multiple clips. Our features not only are time-sensitive but also contain both local fine-grained and global semantic information, which are crucial for temporal reasoning in long videos. Moreover, to further evaluate our model, we present IndustryASM, a large-scale long video dataset of industry assembly processes for temporal action segmentation benchmarking and time and motion studies. Our IndustryASM dataset comprises of $4851$ videos with an average video duration of $105$ seconds. It covers in total $47$ diverse industry assembly tasks and includes timestamp and action labels. We convert the labels into chat samples with manually written instructions. Lastly, extensive experiments demonstrate that TemporalVLM achieves superior results over previous methods. 

In summary, our contributions include:
\begin{itemize}
    \item We develop the first time-aware coarse-to-fine encoder, including a time-aware clip encoder (i.e., overlapping sliding video Q-Former) and a BiLSTM. By leveraging the modules, we tackle both fine-grained understanding and temporal reasoning in long videos.
    \item We present IndustryASM, a large-scale long video dataset of manufacturing assembly procedures. IndustryASM can be downloaded at \url{https://retrocausal.ai/research/}.
    \item TemporalVLM outperforms prior works on temporal reasoning and fine-grained understanding. To our best knowledge, this is the first work to blend LSTMs into video LLMs.
\end{itemize}
\section{Related Work}
\label{sec:relatedwork}

\noindent \textbf{Video Large Language Models.}
Video LLMs~\cite{li2023videochat,liu2023video,luo2023valley,maaz2023video,zhang2023video,jin2024chat,song2024moviechat} typically include a pre-trained visual encoder to extract visual features, a projection layer to map visual features into the text latent space of LLMs, and a pre-trained LLM for generating responses. They mostly differ in the visual encoder. VideoChat~\cite{li2023videochat} extracts frame features via a visual transformer~\cite{sharir2021image} and employs a query transformer (Q-Former)~\cite{li2023blip} to aggregate frame features into video features, while a video Q-Former is further included for temporal modeling in Video-LLaMA~\cite{zhang2023video}. The above methods usually map the video into a fixed number of tokens, yielding degrading performance with long videos, while encoding frames and timestamps separately and hence struggling with temporal reasoning. To address these drawbacks, methods with temporal reasoning capabilities are introduced, e.g., TimeChat~\cite{ren2024timechat} presents a sliding video Q-Former to handle various video lengths and a time-aware frame encoder to jointly encode frames and timestamps. These methods often consider the video as a single clip~\cite{luo2023valley,maaz2023video,li2023videochat,zhang2023video} and aggregate tokens via pooling~\cite{luo2023valley,maaz2023video} and query aggregation~\cite{li2023videochat,zhang2023video}, overlooking fine-grained details. In this work, we divide the video into multiple clips and propose a time-aware clip encoder for capturing fused local features. Also, we integrate a BiLSTM module for aggregating global features.

\noindent \textbf{Video Temporal Reasoning.}
Temporal reasoning plays an important role in video understanding tasks, e.g., dense video captioning~\cite{wang2021end,zhu2022end,yang2023vid2seq}, temporal video grounding~\cite{wang2022negative,luo2023towards}, video highlight detection~\cite{lei2021detecting,moon2023query}, and temporal action segmentation~\cite{li2020ms,yi2021asformer,hyder2024action,lu2024fact}. Prior works~\cite{yang2023vid2seq,luo2023towards,moon2023query,lu2024fact} often focus on designing separate models for tackling individual tasks, while LLM-based models~\cite{qian2024momentor,huang2024vtimellm,huang2024lita,ren2024timechat} capable of handling multiple tasks have emerged recently. Our TemporalVLM model belongs to the second group.

\noindent \textbf{Long Video Understanding.}
Challenges in long video understanding include complex spatial-temporal relationships and redundant information. Long video understanding methods have been developed via efficient architectures~\cite{donahue2015long,kondratyuk2021movinets}, temporal pooling/aggregation~\cite{sener2020temporal,wu2021towards}, and clip selection~\cite{ghodrati2021frameexit,gowda2021smart}. For vision-language understanding tasks, methods based on temporal alignment~\cite{han2022temporal,buch2022revisiting} and memory~\cite{wu2022memvit,Zhao_2023_CVPR} have been introduced. Recently, LongVLM~\cite{weng2024longvlm} divides the video into clips and employs a merging module for extracting local features and pooling operation for computing global features. It does not utilize timestamps, which are crucial for temporal reasoning. Our TemporalVLM model explicitly utilizes timestamps via a time-aware clip encoder. Moreover, we employ a learnable BiLSTM module for aggregating global features.

\noindent \textbf{Procedural Activity Datasets.}
Existing datasets usually focus on cooking activities, e.g., Epic-Kitchens~\cite{damen2018scaling}, are curated from online sources and hence produced with multiple shots, e.g., COIN~\cite{tang2019coin}, and are recorded with toy objects and lab environments, e.g., Assembly101~\cite{sener2022assembly101}. Our IndustryASM dataset focuses on manufacturing assembly processes, is recorded on factory floors, and is labeled by industrial engineers, thereby capturing procedural activities in realistic and practical environments that current datasets have not addressed.
\section{TemporalVLM}
\label{sec:method}

\begin{figure*}[t]
	\centering
		\includegraphics[width=0.9\linewidth, trim = 0mm 0mm 0mm 0mm, clip]{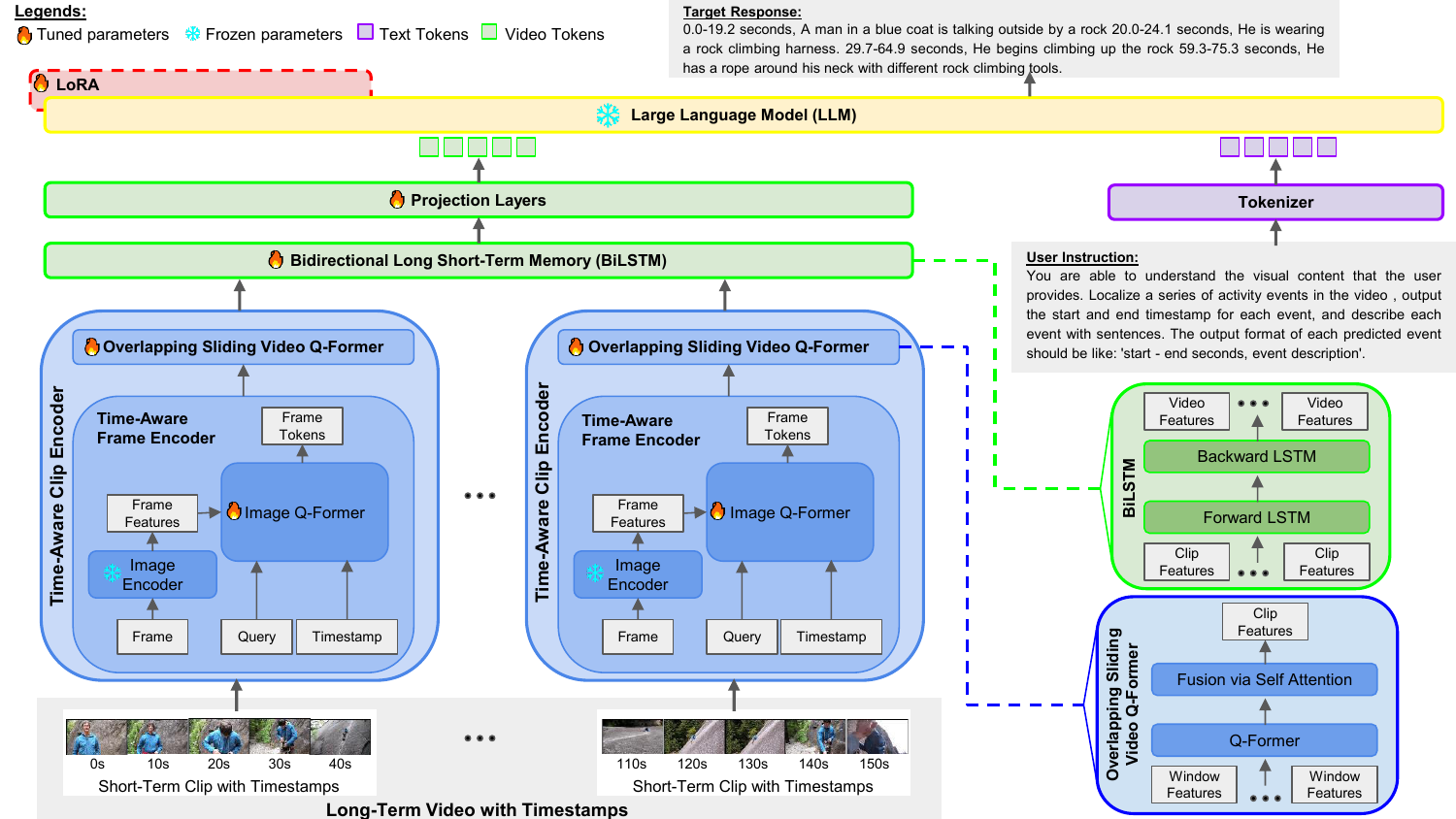}
	\caption{TemporalVLM includes two novel components: a time-aware clip encoder for extracting time-aware fine-grained cues and a BiLSTM module for capturing long-range temporal dependencies.}
	\label{fig:method}
\end{figure*}

An overview of TemporalVLM is shown in Fig.~\ref{fig:method}.

\subsection{Short-Term Temporal Reasoning}
Video LLMs often consider an input video as a single clip and sample a fixed number of frames $N^v_f$ from the video, e.g., $N^v_f = 96$ frames in TimeChat~\cite{ren2024timechat}. Moreover, prior works usually use query aggregation, e.g., TimeChat~\cite{ren2024timechat}, or pooling operation, e.g., Video-ChatGPT~\cite{maaz2023video}, to aggregate video tokens. Thus, they perform well with short videos but miss out fine-grained information for long videos.

We introduce a time-aware clip encoder for extracting fused local fine-grained cues within a clip. We first divide the long-term input video into $C = 6$ short-term clips and sample $N^c_f = 96$ from each clip. Sampled frames of a clip along with timestamps are then passed to our time-aware clip encoder for jointly encoding frame contents and timestamps, yielding fused time-aware local features. Particularly, the time-aware frame encoder uses a pre-trained image encoder~\cite{sun2023eva} to obtain frame features, which are jointly encoded with timestamps via an image Q-Former~\cite{dai2023instructblipgeneralpurposevisionlanguagemodels}, yielding time-aware frame features $\mathbf{f}_t$. 

\noindent \textbf{Overlapping Sliding Video Q-Former.} 
Frame features $\mathbf{f}_t$ are fed to a video Q-Former ~\cite{ren2024timechat} across \textbf{overlapping} windows $\mathbf{W}_i$ of size $q$ = 32 frames and overlap $o$ = 16 frames. Window-wise outputs $\mathbf{V}_i$ are concatenated into $\mathbf{S}$ as:
\begin{align}
    \mathbf{V}_i &= \text{Video Q-Former}(\mathbf{W}_i),\\
    \mathbf{S} &= [\mathbf{V}_1, \mathbf{V}_2, \dots, \mathbf{V}_W], 
\label{eq:v_concat}
\end{align}
with $W$ denoting the number of overlapping windows within a clip.
We propose a \textbf{fusion} module to align diverse temporal cues from redundant boundary tokens in $\mathbf{S}$. Specifically, we apply multi-headed self-attention on $\mathbf{S}$, yielding  $\mathbf{C}$ which fuses local contexts across multiple windows into a single context-aware embedding as:
\begin{align}
    \mathbf{C}^{(h)} &= \text{SoftMax} \left( \frac{\mathbf{Q}^{(h)} {\mathbf{K}^{(h)}}}{\sqrt{d}} \right) \mathbf{V}^{(h)}, \\
    \mathbf{C} &= [\mathbf{C}^{(1)}, \mathbf{C}^{(2)}, \dots, \mathbf{C}^{(H)}] \mathbf{W}^O.
\label{eq:c_concat}
\end{align}
Here, we first project $\mathbf{S}$ into $\mathbf{Q}$, $\mathbf{K}$, $\mathbf{V}$ for each attention head $h$. At each head $h$, $\mathbf{C}^{(h)}$ represents the dot product attention between queries $\mathbf{Q}^{(h)}$ and keys $\mathbf{K}^{(h)}$ scaled over dimension $d$ followed by softmax weighted aggregation over $\mathbf{V}^{(h)}$. $\mathbf{C}$ contains aggregated outputs from all heads with final projection $\mathbf{W}^{O}$ applied. 
Using overlapping windows leads to spatially redundant time-aware tokens in $\mathbf{S}$, each with different local window contexts. This allows $\mathbf{C}$ to yield rich time-aware clip-level information by leveraging diverse temporal views from across overlapping windows. TimeChat's~\cite{ren2024timechat} video Q-Former uses non-overlapping windows and does not perform fusion, yielding inferior results as shown in Tab.~\ref{table:ablation_overlaping}.

\subsection{Long-Term Temporal Reasoning}
Our time-sensitive clip encoder is applied on each short-term clip to obtain fused time-aware local features, which capture useful fine-grained cues for temporal reasoning within the clip. However, they are not effective for temporal reasoning over the long-term video, which requires the ability to capture long-range temporal dependencies. 

\noindent \textbf{Bidirectional Long Short-Term Memory.}
We introduce a BiLSTM module for computing global features across multiple clips. We first concatenate time-aware local features extracted from clips in the temporal order the clips appear in the video. We then pass the sequence of local features to a BiLSTM for aggregating global features. Our BiLSTM follows a standard architecture, including two LSTM networks: one processes the sequence in the original order (forward) and another processes the sequence in the reverse order (backward) as:
\begin{align}
    \mathbf{h}^f_t = \text{LSTM}(\mathbf{h}^f_{t-1}, \mathbf{c}_t),\\
    \mathbf{h}^b_t = \text{LSTM}(\mathbf{h}^b_{t+1}, \mathbf{c}_t),
\end{align}
where $\mathbf{h}^f_t$ and $\mathbf{h}^b_t$ denote the hidden states at time step $t$ of the forward and backward LSTMs respectively and $\mathbf{c}_t$ is the input at time step $t$. The final output $\mathbf{h}_t$ at time step $t$ is obtained by concatenating the outputs of the forward and backward LSTMs as $\mathbf{h}_t = [\mathbf{h}^f_t, \mathbf{h}^b_t]$. Please refer to the textbook~\cite{learning2016ian} for a detailed description. Our BiLSTM module utilizes both past and future information and is capable of capturing long-range temporal relationships in both forward and backward directions. As observed in Tab.~\ref{table:ablation_model}, BiLSTM outperforms various alternatives, including average pooling, linear layer, LSTM, and transformer~\cite{vaswani2017attention}. Unlike LongVLM~\cite{weng2024longvlm}, which applies pooling to obtain global features, our BiLSTM module has learnable parameters and achieves better results. Advanced recurring models, e.g., state space~\cite{gu2021efficiently} and Mamba~\cite{gu2024mamba}, may further improve the performance, which remains our future work.

The video tokens output by our BiLSTM module are of size $(C\times N^c_f,~2\times N_{V})$, while the LLM requires an input size of $(N^c_f,~N_{LLM})$. $N_{V}$ and $N_{LLM}$ are the dimensions of the video tokens and the LLM latent space respectively, and $2$ is to account for the outputs of both forward and backward passes of BiLSTM. Thus, we pass the video tokens output by BiLSTM through projection layers to match the input dimensions required by the LLM.

\subsection{Large Language Model}
The LLM takes as input the video tokens $\mathbf{X}^v$, query tokens $\mathbf{X}^q$, and generates responses $\mathbf{X}^r$ to users. Video LLMs typically employ a two-stage training: leveraging large-scale image/video-text pairs for vision-language alignment to pre-train the model, and utilizing instruction data to fine-tune the pre-trained model. Following TimeChat~\cite{ren2024timechat}, we use the checkpoint of the LLaMA-2 7B model~\cite{touvron2023llama2} and perform instruction tuning only. We optimize the below loss:
\begin{align}
    \mathcal{L} = - \log P_{\mathbf{\psi}} (\mathbf{X}^r | \mathbf{X}^v, \mathbf{X}^q ) \\
    = - \sum^{L}_{i=1} \log P_{\mathbf{\psi}} (x_i | \mathbf{X}^v, \mathbf{X}^q, \mathbf{X}^r_{<i} ).
\end{align}
Here, $\mathbf{\psi}$ is the learnable parameters of TemporalVLM, $L$ is the response length, $x_i$ is the current predicted token, and $\mathbf{X}^r_{<i}$ is the prior tokens appearing before $x_i$ in the response.

\section{IndustryASM}
\label{sec:industryasm}

\begin{figure*}[t]
	\centering
		\includegraphics[width=0.9\linewidth, trim = 0mm 100mm 0mm 0mm, clip]{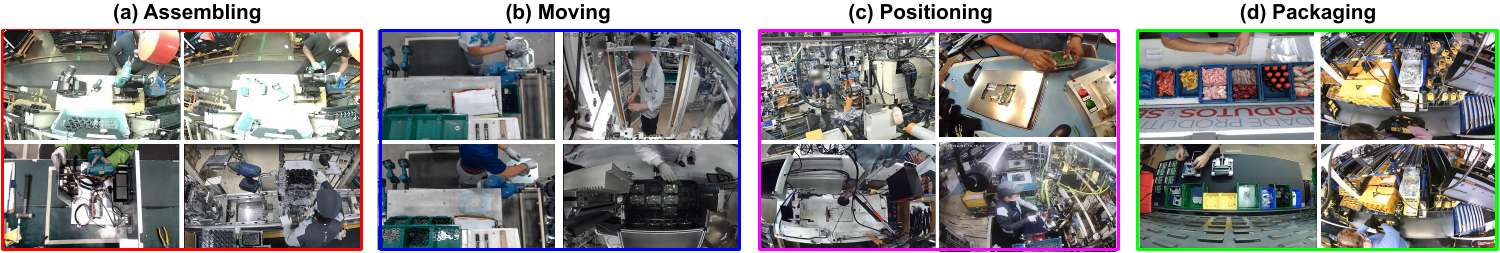}
	\caption{Example IndustryASM videos with different camera viewpoints, actors, backgrounds, and activities.}
	\label{fig:industryasm}
\end{figure*}

\subsection{Dataset Statistics}
IndustryASM includes 4851 videos with each lasting 105 seconds on average, yielding 142 hours as the total dataset duration. In addition, it consists of 47 various industry assembly procedures with each having 12 steps on average. Example videos are shown in Fig.~\ref{fig:industryasm}. Due to space limits, we provide additional details in our supplementary material.

\subsection{Annotation Process}
Our videos are annotated by industrial engineers into framewise action segments. The action labels contain the same step naming conventions as used on factory floors during manufacturing processes. To ensure quality, each video is annotated by two labelers, i.e., one obtains the labels for the video, while another checks the labels. In cases of conflicts, both labelers discuss to address them. Overall, around 8\% of the videos have conflicts and require fixing, yielding an agreement rate of 92\% between labelers. Given the framewise annotations we first convert them to timestamps by using the frame rate of the video. We then manually write the instructions designed for the action segmentation task to ensure quality and generate the answers by using the timestamps and action labels. Examples are included in our supplementary material.

\section{Experiments}
\label{sec:experiments}

\noindent \textbf{Tasks and Datasets.}
We evaluate TemporalVLM on dense video captioning, temporal video grounding, video highlight detection, and temporal action segmentation. For fair comparisons with TimeChat~\cite{ren2024timechat}, we fine-tune our model on a subset\footnote{We could not download YT-Temporal~\cite{zellers2022merlot} due to its large size and the restricted number of downloads.} of TimeIT~\cite{ren2024timechat} and Valley~\cite{luo2023valley}. The combined dataset contains 142K long videos with diverse activities. We follow TimeChat to use 6 instructions (generated by GPT-4~\cite{achiam2023gpt}) per task and convert video annotations into chat samples. We evaluate on YouCook2~\cite{zhou2018towards} for dense video captioning, Charades-STA~\cite{gao2017tall} for temporal video grounding, QVHighlights~\cite{lei2021detecting} for video highlight detection, and IndustryASM for temporal action segmentation. Note that only Tab.~\ref{table:assembly} uses \textbf{our} dataset, all the remaining evaluations use \textbf{existing} datasets. Following prior works, we report the results from a single run.

\noindent \textbf{Dataset Splits}
For dense video captioning, YouCook2~\cite{zhou2018towards} is divided into 1333 videos for fine-tuning out of a total of 1790 videos, while 457 videos are used for evaluation. For the temporal video grounding task, we divide the 16124 video-query pairs of Charades-STA~\cite{gao2017tall} into 12404 fine-tuning pairs and 3720 evaluation pairs. For video highlight detection, 6858 videos are used for fine-tuning and 1463 are used for evaluation from the QVHighlights~\cite{lei2021detecting} dataset. These are the default splits provided by each dataset. Finally, we divide our IndustryASM dataset into 3896 videos for fine-tuning and 955 videos for evaluation.

\noindent \textbf{Implementation Details.}
We follow TimeChat~\cite{ren2024timechat} to use LLaMA-2 7B~\cite{touvron2023llama2} as the LLM, and ViT-G/14 from EVA-CLIP~\cite{sun2023eva} as the image encoder. We take InstructBLIP's~\cite{dai2023instructblipgeneralpurposevisionlanguagemodels} checkpoints to initialize the image Q-Former, and Video-LLaMA’s~\cite{zhang2023video} checkpoints to initialize the video Q-Former. We employ pyTorch's~\cite{imambi2021pytorch} multiheaded attention in the fusion module. The BiLSTM module includes forward and backward LSTMs with one hidden layer each. The BiLSTM and projection layers are initialized randomly, while the LLM and image encoder remain frozen throughout fine-tuning. The LLM is fine-tuned using LoRA~\cite{hu2023lora}. We fine-tune our model on an 8-A100 (80 GB) machine. Please see our supplementary material for more details.

\noindent \textbf{Competing Methods.} 
We compare our model against VideoChat-Embed~\cite{li2023videochat}, Video-LLaMA~\cite{zhang2023video}, Valley~\cite{luo2023valley}, TimeChat~\cite{ren2024timechat}, and LongVLM~\cite{weng2024longvlm}. Since LongVLM is not trained on TimeIT~\cite{ren2024timechat} and Valley~\cite{luo2023valley} and it is a multi-model pipeline (TimeChat and TemporalVLM are end-to-end), we replace our time-aware clip encoder and BiLSTM with their local and global feature extraction modules respectively while keeping the same image encoder and LLM, yielding our reimplemented LongVLM for evaluation. Nevertheless, we compare with the original LongVLM on general video understanding in our supplementary material.

\noindent \textbf{Evaluation Metrics.} 
We use the below metrics:
\begin{itemize}
    \item Dense video captioning: We use SODA\textunderscore c~\cite{fujita2020soda} for story evaluation, F1 Score for event localization, and CIDEr~\cite{vedantam2015cider} for measuring caption quality.
    \item Temporal video grounding: We use R@1\textsubscript{(IoU=$\alpha$)}, which is the percentage of predictions with an intersection over union greater than $\alpha$ compared to the ground truth.
    \item Video highlight detection: We use mean average precision (mAP) and HIT@1 (IoU= 0.75) to evaluate saliency scores of key moments.
    \item Temporal action segmentation: We use framewise accuracy, F1 scores at overlapping thresholds $\{0.1, 0.25, 0.5\}$, and Edit distance.
\end{itemize}

\subsection{Zero-Shot Results}
\begin{table*}[t!]
\centering
\small
\begin{tabular}{@{}lclll|ll|ll@{}} 
\toprule
\multirow{3}{*}{\textbf{Model}} & \multirow{3}{*} & \multicolumn{3}{c|}{\textbf{Dense Captioning}} & \multicolumn{2}{c|}{\textbf{Highlight Detection}} & \multicolumn{2}{c}{\textbf{Temporal Grounding}} \\
                 &  & \multicolumn{3}{c|}{\textbf{YouCook2}}                 & \multicolumn{2}{c|}{\textbf{QVHighlights}}               & \multicolumn{2}{c}{\textbf{Charades-STA}}              \\ \cmidrule(l){3-9} 
                 &  & \textbf{SODA\_c}   & \textbf{CIDEr}        & \textbf{F1}          & \textbf{mAP}                  & \textbf{HIT@1}        & 
                 \textbf{R@1 \textsubscript{(IoU=0.5)}}             
                 & 
                 \textbf{R@1 \textsubscript{(IoU=0.7)}}   
                 \\ \midrule

Valley\textsubscript{~\cite{luo2023valley}}&   &  0.1 &   0.0  & 1.5  &  10.9 &   15.2 &  4.7 &    1.6 \\
Video-LLaMA\textsubscript{~\cite{zhang2023video}}&  & 0.0 & 0.0 & 0.1 & 11.3 & 15.6 & 2.7 & 1.2 \\
VideoChat-Embed\textsubscript{~\cite{li2023videochat}}&   &    0.2   &  0.6   &   3.4   &    13.1    &   18.1  &  3.2          &   1.4   \\

LongVLM\textsubscript{~\cite{weng2024longvlm}}&  & \underline{\textit{0.8}}   &  2.5   &   12.3   &    11.1    &   15.0  &  13.9          &   6.1   \\

TimeChat \textsubscript{~\cite{ren2024timechat}}&  &  \textbf{1.2} &  \underline{\textit{3.4}} & \underline{\textit{12.6}}  &   \underline{\textit{14.5}}  &  \underline{\textit{23.9}}  &   \underline{\textit{27.9}}   &   \underline{\textit{12.3}} 
\\
TemporalVLM& &  \textbf{1.2} &  \textbf{3.7} & \textbf{13.1}  &   \textbf{16.4}  &  \textbf{31.3}  &   \textbf{30.1}   & 
\textbf{13.2}\\

\bottomrule
\end{tabular}

\caption{Zero-short results. We test on YouCook2 for dense video captioning, QVHighlights for video highlight detection, Charades-STA for temporal video grounding. Best results are in \textbf{bold}, second best ones are \underline{\textit{underlined}}.}
\label{table:zeroshot}
\end{table*}

\noindent \textbf{Dense Video Captioning Results.} Tab.~\ref{table:zeroshot} shows that TemporalVLM achieves the best results across all metrics. For example, we outperform TimeChat~\cite{ren2024timechat} by \textbf{+0.3} on CIDEr and \textbf{+0.5} on F1 Score. This is likely because TimeChat treats the entire video as a single clip and aggregates video tokens via query aggregation, overlooking fine-grained information. In contrast, our extracted features, which capture both local fine-grained cues from each clip and global temporal dependencies across multiple clips, are effective for caption generation and event localization. Our approach also obtains superior results than LongVLM~\cite{weng2024longvlm} across SODA\_c \textbf{(+0.4)}, CIDEr (\textbf{+1.2}), and F1 Score (\textbf{+0.8}), which confirms the importance of our time-aware clip encoder and BiLSTM for dense video captioning. 

\noindent \textbf{Video Highlight Detection Results.} From Tab.~\ref{table:zeroshot}, TemporalVLM obtains the best performance on both mAP and HIT@1. It surpasses TimeChat~\cite{ren2024timechat} by \textbf{+1.9} in mAP and \textbf{+7.4} in HIT@1, while outperforming LongVLM~\cite{weng2024longvlm} by \textbf{+5.3} at mAP and by \textbf{+16.3} at HIT@1. These results suggest the effectiveness of our model for video highlight detection. By aggregating global features from time-aware local features, it identifies the most relevant timestamps and assigns them saliency scores based on their relative importance within the video for a given query. Though LongVLM extracts multi-level features, the lack of a time-aware encoder hinders its results. 

\noindent \textbf{Temporal Video Grounding Results.} It is evident from Tab.~\ref{table:zeroshot} that TemporalVLM achieves the best performance across R@1\textsubscript{(IoU=0.5)} and R@1\textsubscript{(IoU=0.7)}. It outperforms TimeChat~\cite{ren2024timechat} by \textbf{+2.2} and \textbf{+0.9} on R@1\textsubscript{(IoU=0.5)} and R@1\textsubscript{(IoU=0.7)} respectively, and LongVLM~\cite{weng2024longvlm} by \textbf{+16.2} and \textbf{+7.1} on R@1\textsubscript{(IoU=0.5)} and R@1\textsubscript{(IoU=0.7)} respectively. The results validate the benefits of our time-aware and multi-level features for temporal video grounding. Our clip encoder extracts fine-grained cues for temporal reasoning within each clip via fusion, which enables our BiLSTM to capture long-range temporal relationships across clips to detect the given event effectively. The lack of a time-aware encoder affects LongVLM, while TimeChat suffers from a lack of fine-grained cues.

\subsection{Supervised Results}

\noindent \textbf{Dense Video Captioning, Video Highlight Detection, and Temporal Video Grounding Results.} Tab.~\ref{table:finetune} presents the results in the supervised setting. TemporalVLM performs the best across all tasks and metrics. As compared to the zero-shot results in Tab.~\ref{table:zeroshot}, we see an increase in performance, e.g., \textbf{+24.3} on R@1\textsubscript{(IoU=0.5)} in temporal video grounding and \textbf{+9.5} on CIDEr in dense video captioning. TemporalVLM also outperforms more recent long video understanding models, such as Gen-S~\cite{yao2025generative}, by \textbf{+15.7} and \textbf{+13.8} at R@1\textsubscript{(IoU=0.5)} and R@1\textsubscript{(IoU=0.7)}, respectively. Furthermore, TemporalVLM performs better than newer temporal reasoning methods, i.e., NumPro-FT~\cite{wu2025number} and LLAVA-ST~\cite{li2025llava}. Our model shows an improvement of \textbf{+12.4} and \textbf{+8.5} compared to NumPro-FT and \textbf{+9.6} and \textbf{+5.6} compared to LLAVA-ST at R@1\textsubscript{(IoU=0.5)} and R@1\textsubscript{(IoU=0.7)}, respectively. However, we acknowledge that task-specific models, e.g., MMN~\cite{wang2022negative} and QD-DETR~\cite{moon2023query}, may outperform generalist models, e.g., TimeChat~\cite{ren2024timechat} and TemporalVLM, due to their task-specific designs. Due to space limits, we provide a comparison with task-specific models in our supplementary material.
\begin{table*}[t!]
\centering
\small
\begin{tabular}{@{}lclll|ll|ll@{}} 
\toprule
\multirow{3}{*}{\textbf{Model}} & \multirow{3}{*} & \multicolumn{3}{c|}{\textbf{Dense Captioning}} & \multicolumn{2}{c|}{\textbf{Highlight Detection}} & \multicolumn{2}{c}{\textbf{Temporal Grounding}} \\
                 &  & \multicolumn{3}{c|}{\textbf{YouCook2}}                 & \multicolumn{2}{c|}{\textbf{QVHighlights}}               & \multicolumn{2}{c}{\textbf{Charades-STA}}              \\ \cmidrule(l){3-9} 
                 &  & \textbf{SODA\_c}   & \textbf{CIDEr}        & \textbf{F1}          & \textbf{mAP}                  & \textbf{HIT@1}        & 
                 \textbf{R@1 \textsubscript{(IoU=0.5)}}             
                 & 
                 \textbf{R@1 \textsubscript{(IoU=0.7)}}   
                 \\ \midrule

LongVLM\textsubscript{~\cite{weng2024longvlm}} & & 2.3   &  8.1   &   16.9   &    16.0    &   22.5  &  27.2          &   11.9   \\

TimeChat\textsubscript{~\cite{ren2024timechat}} & &  \underline{\textit{3.1}} &  \underline{\textit{10.3}} & \underline{\textit{19.5}}  &   \underline{\textit{21.7}}  &  \underline{\textit{37.9}}  &   \underline{\textit{46.7}}   &   \underline{\textit{23.7}} 
\\

Gen-S\textsubscript{~\cite{yao2025generative}} & & -- &  -- & -- &   --  &  -- &  38.7   &  15.2
\\

NumPro-FT\textsubscript{~\cite{wu2025number}} & & -- &  -- & -- &   --  &  -- &  42.0   &  20.6
\\

LLAVA-ST\textsubscript{~\cite{li2025llava}} & & -- &  -- & -- &   --  &  -- &  44.8   &  23.4
\\

TemporalVLM &  & $\textbf{3.4}$ &  
$\textbf{13.2}$ & 
$\textbf{20.0}$  & $\textbf{25.1}$  &  $\textbf{43.0}$  & $\textbf{54.4}$   & $\textbf{29.0}$ \\

\bottomrule
\end{tabular}

\caption{Supervised results. We first fine-tune all models on TimeIT and Valley, and then perform task-specific fine-tuning and evaluation on YouCook2 for dense video captioning, QVHighlights for video highlight detection, and Charades-STA for temporal video grounding. Best results are in \textbf{bold}, while second best ones are \underline{\textit{underlined}}.}
\label{table:finetune}
\end{table*}

\noindent \textbf{Temporal Action Segmentation Results.} We now evaluate on our IndustryASM dataset. Since the activities in IndustryASM significantly differ from those in TimeIT~\cite{ren2024timechat} and Valley~\cite{luo2023valley}, we fine-tune all models on IndustryASM before evaluation. Tab.~\ref{table:assembly} presents the results. TemporalVLM performs the best across all metrics. For example, it outperforms TimeChat~\cite{ren2024timechat} by \textbf{+23.5} and \textbf{+6.9} on Edit and Acc respectively, and LongVLM~\cite{weng2024longvlm} by \textbf{+8.9} and \textbf{+5.3} on Edit and Acc respectively. The results confirm the benefits of our time-aware and multi-level features.
\begin{table}[t!]

\centering
\small
\begin{tabular}{@{}llll@{}}
\toprule
\textbf{Model} &  \textbf{F1@\{10,25,50\}} & \textbf{Edit} & \textbf{Acc} \\ \midrule
LongVLM\textsubscript{~\cite{weng2024longvlm}} & \{\underline{\textit{17.1}}, \underline{\textit{13.3}}, \underline{\textit{7.3}}\} & \underline{\textit{47.3}} & \underline{\textit{47.6}} \\
TimeChat\textsubscript{~\cite{ren2024timechat}} & \{11.1, 8.5, 4.4\} & 32.7 & 46.0 \\
TemporalVLM & \{\textbf{22.3}, \textbf{18.3}, \textbf{11.1}\} & \textbf{56.2} & \textbf{52.9}  \\ 
\bottomrule
\end{tabular}
\caption{Temporal action segmentation results in supervised setting on IndustryASM. All models are fine-tuned on IndustryASM before evaluation. Best results are in \textbf{bold}, while second best ones are \underline{\textit{underlined}}.}

\label{table:assembly}
\end{table}

\noindent \textbf{Multi-Clip Encoder vs. Time-Aware Encoder.} 
Multi-clip encoders in LongVLM~\cite{weng2024longvlm} and TemporalVLM are crucial to capturing fine-grained cues, leading to superior results over TimeChat~\cite{ren2024timechat} on temporal action segmentation in Tab.~\ref{table:assembly}. In contrast, time-aware encoders in TimeChat and TemporalVLM are important to event localization and temporal modeling, yielding better results than LongVLM on dense video captioning, video highlight detection, and temporal video grounding in Tabs.~\ref{table:zeroshot} and~\ref{table:finetune}. TemporalVLM is both multi-clip and time-aware.

\subsection{Ablation Results}
\label{sec:ablation}

We fine-tune several variants of TemporalVLM on YouCook2~\cite{zhou2018towards} to study our overlapping sliding video Q-Former and BiLSTM.

\noindent \textbf{Impacts of Overlapping Sliding Video Q-Former.} Tab.~\ref{table:ablation_overlaping} demonstrates the importance of fusing encoded frames across overlapping temporal windows in our clip encoder. Our model with fusion across overlapping windows achieves the best results, whereas using non-overlapping windows and removing the fusion module as in TimeChat's~\cite{ren2024timechat} design lead to performance drops. The results validate our design.

\begin{table}[t!]
\centering
\small
\setlength{\tabcolsep}{5pt} 
\begin{tabular}{@{}lccc@{}}
\toprule
\textbf{Model} & \textbf{SODA} & \textbf{CIDEr} & \textbf{F1} \\ \midrule
\makecell[l]{No Overlap + No Fusion} & 2.3 & 11.0 & 17.3 \\
\makecell[l]{No Overlap + Fusion} & 2.0 & 9.5 & 16.2 \\
\makecell[l]{Overlap + No Fusion} & \underline{\textit{2.8}} & \underline{\textit{11.7}} & \underline{\textit{18.9}} \\
\makecell[l]{Overlap + Fusion} & \textbf{3.4} & \textbf{13.2} & \textbf{20.0} \\
\bottomrule
\end{tabular}
\caption{Effects of overlapping sliding video Q-Former. Best results are in \textbf{bold}, second best are \underline{\textit{underlined}}.}
\label{table:ablation_overlaping}
\end{table}

\noindent \textbf{Impacts of BiLSTM.} To study the effects of BiLSTM, we replace it with various alternatives and report the results in Tab.~\ref{table:ablation_model}. From Tab.~\ref{table:ablation_model}, BiLSTM outperforms average pooling, which is not trained and employed in prior works~\cite{maaz2023video,weng2024longvlm}, and traditional LSTM, which only performs the forward pass and hence relies only on past cues. In contrast, BiLSTM is learnable and conducts both forward and backward passes to utilize both past and future cues. Moreover, we replace BiLSTM with linear layer to examine the impacts of video Q-Formers~\cite{zhang2023video,ren2024timechat} alone, leading to worse results. This highlights the importance of BiLSTM even when video Q-Formers are being used. Finally, BiLSTM outperforms transformer~\cite{vaswani2017attention}, which we will discuss below.

\begin{table}[t!]

\centering
\small
\begin{tabular}{@{}lllll@{}}
\toprule
\textbf{Model} & \textbf{\# Layers} &  \textbf{SODA\_c}   & \textbf{CIDEr}  & \textbf{F1} \\ \midrule
Average Pooling & 1   & 1.7 & 6.0 & 16.1 \\
Linear Layer &  1   & 2.4 & \underline{\textit{9.5}} & 17.8 \\
Transformer & 2   & 2.3 & 6.8 & 15.6 \\
LSTM & 1   & 2.8 & 9.4 & 18.2 \\
LSTM & 2   & \underline{\textit{2.9}} & 9.3 & \underline{\textit{18.5}} \\
BiLSTM & 2  & \textbf{3.4} & \textbf{13.2} & \textbf{20.0} \\ 
\bottomrule
\end{tabular}
\caption{Effects of BiLSTM module. Best results are in \textbf{bold}, while second best ones are \underline{\textit{underlined}}.}

\label{table:ablation_model}
\end{table}

\noindent \textbf{(Bi)LSTM vs. Transformer.}
The permutation-invariant nature of the self-attention mechanism in transformer~\cite{vaswani2017attention} leads to loss in temporal information of the input sequence. Though using positional encoding alleviates this issue, it is inevitable that some temporal information loss occurs~\cite{zeng2023transformers}. In contrast, (Bi)LSTM processes data one by one and preserves the sequence order, making them better suited for temporal modeling. Similar to~\cite{zeng2023transformers}, we also observe in Tab.~\ref{table:ablation_model} that linear layer outperforms transformer~\cite{vaswani2017attention}.

\subsection{Comparisons with Specialized Models}
\label{sec:supp_specialized_models}

Tab.~\ref{table:supp_specialized} shows that specialized models such as Vid2Seq~\cite{yang2023vid2seq} and QD-DETR~\cite{moon2023query}, whose advantages come from task-specific training data and model designs, outperform generalist models such as TimeChat~\cite{ren2024timechat} and our TemporalVLM model. In particular, the use of 4 loss functions for 200 epochs during training and the utilization of saliency tokens for saliency prediction in QD-DETR lead to superior performance in video highlight detection tasks. In contrast, our TemporalVLM model is trained on a simpler language modeling loss for a smaller number of epochs. Nevertheless, generalist models exhibit better generalization across zero-shot, multi-task, and multi-domain settings.  

\begin{table*}[t]
\small
\centering
\renewcommand{\arraystretch}{1.0}
\begin{tabular}{@{}lclll|ll|ll@{}} 
\toprule
\multirow{3}{*}{\textbf{Model}} & \multirow{3}{*} & \multicolumn{3}{c|}{\textbf{Dense Captioning}} & \multicolumn{2}{c|}{\textbf{Highlight Detection}} & \multicolumn{2}{c}{\textbf{Temporal Grounding}} \\
                 &  & \multicolumn{3}{c|}{\textbf{YouCook2}}                 & \multicolumn{2}{c|}{\textbf{QVHighlights}}               & \multicolumn{2}{c}{\textbf{Charades-STA}}              \\ \cmidrule(l){3-9} 
                 &  & \textbf{SODA\_c}   & \textbf{CIDEr}        & \textbf{F1}          & \textbf{mAP}                  & \textbf{HIT@1}        & 
                 \textbf{R@1 \textsubscript{(IoU=0.5)}}             
                 & 
                 \textbf{R@1 \textsubscript{(IoU=0.7)}}   
                 \\ \midrule

Vid2Seq \textsubscript{\cite{yang2023vid2seq}} & & \textbf{7.9} & \textbf{47.1} & \textbf{27.3} & -- & -- & -- & -- \\
QD-DETR \textsubscript{\cite{moon2023query}} & & -- & -- & -- & \underline{\textit{38.9}} & \underline{\textit{62.4}} & -- & -- \\
QD-DETR w/ Audio \textsubscript{\cite{moon2023query}} & & -- & -- & -- & \textbf{39.0} & \textbf{62.9} & -- & -- \\
MMN \textsubscript{\cite{wang2022negative}} & & -- & -- & -- & -- & -- & 50.5 & \underline{\textit{29.7}} \\
VDI \textsubscript{\cite{luo2023towards}} & & -- & -- & -- & -- & -- & \underline{\textit{52.3}} & \textbf{31.4} \\
\midrule
Timechat \textsubscript{\cite{ren2024timechat}}  & & 3.1 & 10.3 & 19.5 & 21.7 & 37.9 & 46.7 & 23.7 \\
TemporalVLM & & \underline{\textit{3.4}} & \underline{\textit{13.2}} & \underline{\textit{20.0}} & 25.1 & 43.0 & \textbf{54.4} & 29.0 \\
\bottomrule
\end{tabular}
\caption{Comparisons with specialized models. Best results are in \textbf{bold}, while second best ones are \underline{\textit{underlined}}.}
\label{table:supp_specialized}
\end{table*}

\subsection{Qualitative Results}

\begin{figure}[t]
	\centering
		\includegraphics[width=0.85\linewidth, trim = 0mm 0mm 0mm 0mm, clip]{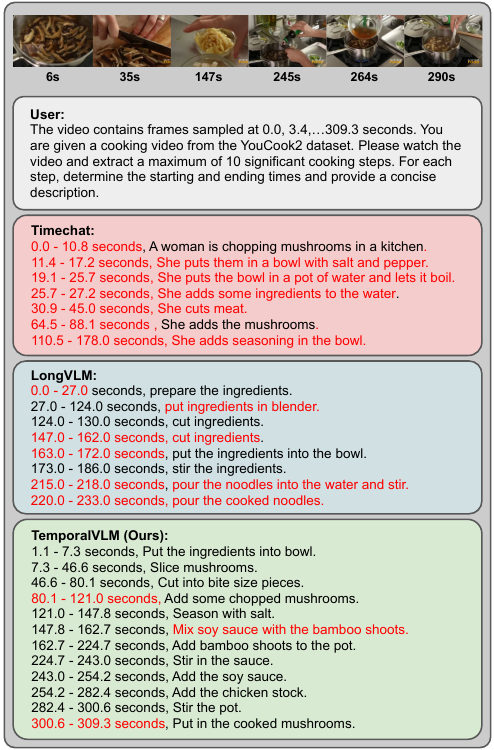}
	\caption{Dense video captioning in zero-shot setting on YouCook2. \textbf{\textcolor{red}{Red}} denotes inaccuracies.}
	\label{fig:qualitative_youcook}
\end{figure}

\begin{figure}[t]
	\centering
		\includegraphics[width=0.9\linewidth, trim = 0mm 0mm 0mm 0mm, clip]{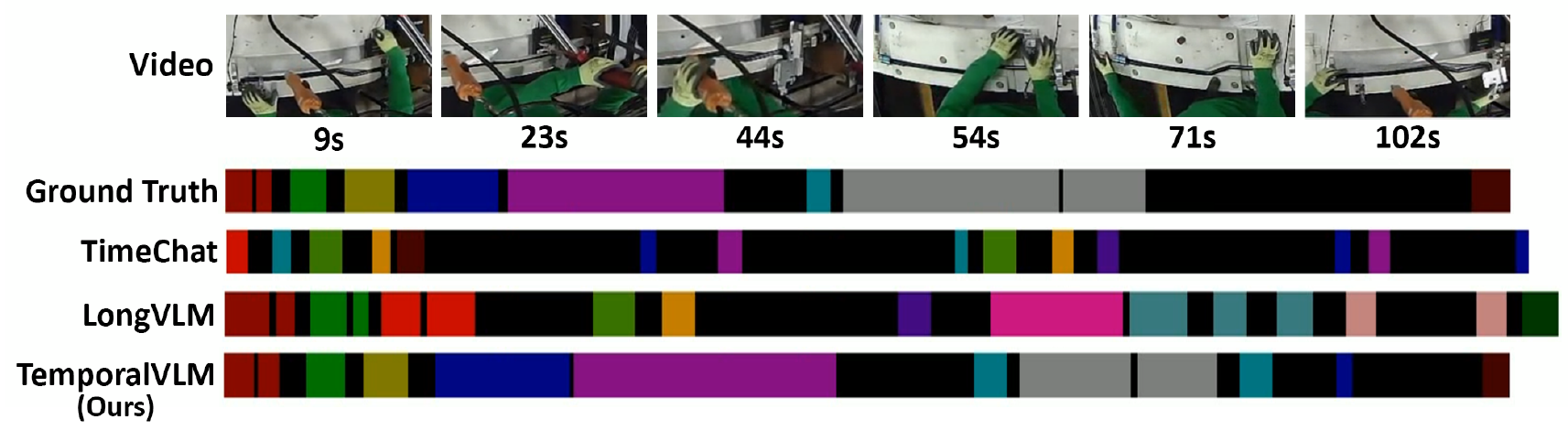}
	\caption{Temporal action segmentation in supervised setting on IndustryASM. \textbf{Black} denotes background.}
	\label{fig:qualitative_asm}
\end{figure}

\noindent \textbf{Dense Video Captioning Results.} Fig.~\ref{fig:qualitative_youcook} shows example results in the zero-shot setting on a YouCook2~\cite{zhou2018towards} video. TimeChat~\cite{ren2024timechat} fails to predict any correct timestamps and tends to hallucinate. LongVLM~\cite{weng2024longvlm} yields better results but still hallucinates objects not present in the video and produces inaccurate captions. In contrast, TemporalVLM produces much more accurate timestamps and captions. Also, it is the only model that provides captions till the end of the video, showing the effectiveness for long video understanding.

\noindent \textbf{Temporal Action Segmentation Results.} Fig.~\ref{fig:qualitative_asm} presents example results in the supervised setting on an IndustryASM video. TimeChat~\cite{ren2024timechat} fails to segment the video and hallucinates actions not present in the video. LongVLM~\cite{weng2024longvlm} fares better in terms of the actions detected in the video but struggles with predicting the action segments and hallucinations further into the video. In contrast, TemporalVLM predicts the action segments notably closer to the ground truth.

\section{Discussion}

\noindent \textbf{Novelty of TemporalVLM.} Existing frameworks either lack explicit time-aware modeling (Video-ChatGPT~\cite{maaz2023video}, LongVLM~\cite{weng2024longvlm}) or do not adopt a coarse-to-fine design (Video-ChatGPT, TimeChat~\cite{ren2024timechat}), and typically aggregate global information via pooling (Video-ChatGPT, LongVLM) or query-based aggregation (TimeChat). In contrast, our time-aware coarse-to-fine encoder explicitly models temporal information, with (i) a time-aware clip encoder extracting fused time-aware local features and (ii) a BiLSTM aggregating these features to capture long-range temporal dependencies. This enables fine-grained video understanding and temporal reasoning within a unified framework, yielding consistent improvements.

\noindent \textbf{Generalizability of IndustryASM.} While the focus of IndustryASM is on industrial assembly, the underlying tasks, i.e., multi-step procedure understanding, temporal dependency reasoning, and long-range event modeling, are common across many long-video domains, including instructional, maintenance, and tutorial videos. Its complex temporal structures, long durations, and multi-step workflows make it a challenging benchmark. Consequently, despite domain specificity, IndustryASM captures the general characteristics of procedural long videos and provides a valuable benchmark for evaluating temporal reasoning models beyond industrial settings.
\section{Conclusion}
\label{sec:conclusion}

We propose TemporalVLM for temporal reasoning and fine-grained understanding in long videos. Our approach includes a time-aware clip encoder for extracting fused time-aware local features and a BiLSTM for global feature aggregation. Extracted features are time-sensitive and contain both local and global cues. Moreover, we present IndustryASM. Lastly, extensive experiments show our superior results over prior works. To our best knowledge, this is the first work to blend LSTMs into video LLMs. Our future work will explore advanced recurring models~\cite{gu2021efficiently,gu2024mamba}.

\bibliography{references}

\newpage

\appendix
\section*{Supplementary Material}

In this supplementary material, we first provide the additional details of our IndustryASM dataset (including dataset statistics, annotation process, maintenance and release plans) and our TemporalVLM implementation in Secs.~\ref{sec:supp_industryasm} and~\ref{sec:supp_implementation} respectively. Next, we demonstrate the effects of number of training epochs in Sec.~\ref{sec:supp_no_of_epochs}. We then evaluate our model for general video understanding in Sec.~\ref{sec:supp_benchmark}. Furthermore, Sec.~\ref{sec:supp_model_size} compares the sizes of our model against TimeChat~\cite{ren2024timechat} and LongVLM~\cite{weng2024longvlm}, while we report the run time comparisons and FLOP comparisons in Secs.~\ref{sec:supp_inference_time} and \ref{sec:flops_comparison} respectively to evaluate the efficiency of our model. Moreover, Sec.~\ref{sec:supp_no_of_clips} assesses the impacts of number of clips, while we study the effects of employing a more advanced LLM in our model in Sec.~\ref{sec:supp_llm_models}. We then present several qualitative results in Sec.~\ref{sec:supp_qualitative}, including dense video captioning, video highlight detection, temporal video grounding, and more importantly, generalization results. Finally, we discuss the limitations and societal impacts of our work in Secs.~\ref{sec:supp_limit} and \ref{sec:supp_societal} respectively.

\section{IndustryASM Details}
\label{sec:supp_industryasm}

\subsection{Dataset Statistics}
Our IndustryASM dataset comprises of 4851 videos in total and the average video duration is 105 seconds. Therefore, the total dataset duration is 142 hours. These videos are distributed among 47 industry assembly processes or datasets, ranging from automotive manufacturing, electronic device manufacturing, medical device manufacturing to heating, ventilation, and air conditioning (HVAC) manufacturing. Participants in our dataset are salaried factory workers based in the US. Informed consent is obtained from all participants in accordance with standard operating procedures prior to data collection. As an additional privacy safeguard, all faces and other personally identifiable information (PII) are blurred or otherwise anonymized in the recorded data. Each assembly process or dataset involves 12 steps or actions on average. We first  categorize each step into one of the following classes:
\begin{itemize}
	\item \textbf{Moving:} In this category, the worker(s) transfer parts or subassemblies within or between workstations, relocating items from one position to another within the workspace.
	\item \textbf{Assembling:} The worker(s) build or complete subassemblies by adding components or combining parts to form a complete unit.
	\item \textbf{Positioning:} This involves the worker(s) placing parts onto a subassembly or mounting a subassembly in a designated position within the workstation.
	\item \textbf{Packaging:} In this category, the worker(s) place subassemblies or components into boxes or other forms of packaging materials.
\end{itemize}

\begin{figure}[t]
	\centering
		\includegraphics[width=\linewidth, trim = 0mm 0mm 0mm 0mm, clip]{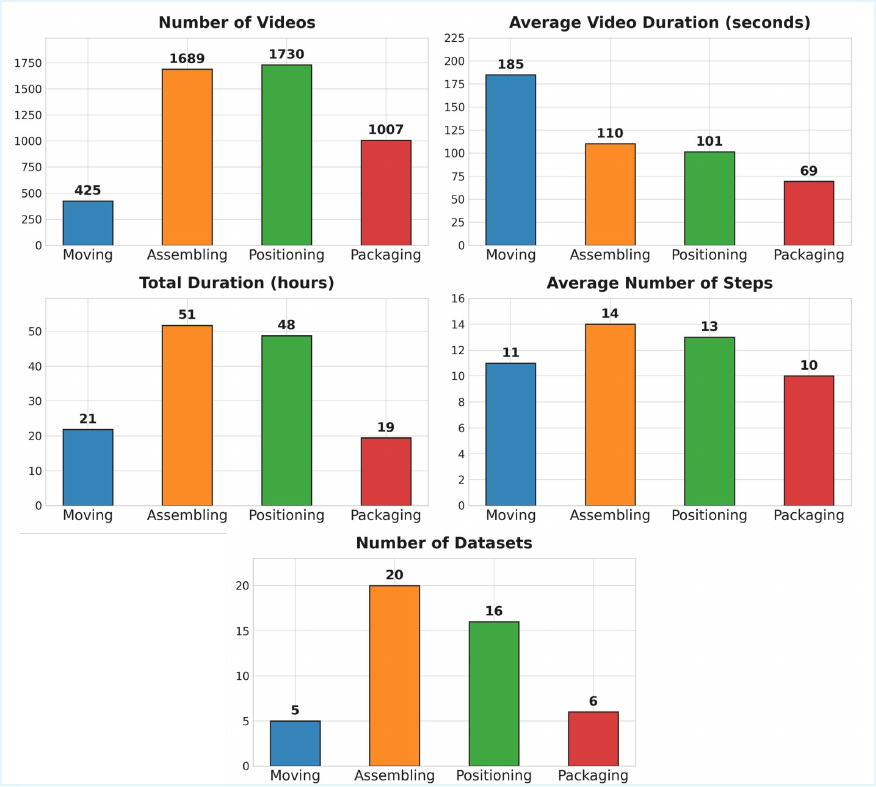}
	\caption{Dataset statistics for our IndustryASM dataset.}
	\label{fig:supp_dataset_statistics}
\end{figure}

Next, we further classify each assembly process or dataset into one of the above categories which appears the most among all of its steps. For example, the Air\_Cleaner dataset (P06) includes 10 steps (namely, 4 moving steps, 3 assembling steps, and 3 positioning steps), and hence it is classified as ``Moving'' which appears the most among its 10 steps. We provide the statistics for each category of the IndustryASM dataset in Fig.~\ref{fig:supp_dataset_statistics}.

\begin{figure*}[t]
	\centering
		\includegraphics[width=\linewidth, trim = 0mm 100mm 0mm 0mm, clip]{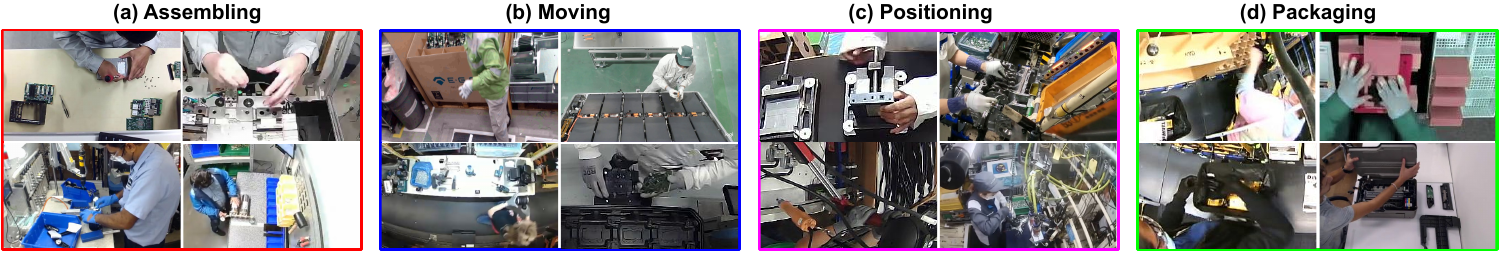}
	\caption{Example videos with different camera viewpoints, actors, backgrounds, and activities from our IndustryASM dataset.}
	\label{fig:supp_industryasm_example}
\end{figure*}

Finally, examples of these 4 categories are presented in Fig.~3 of the main paper. In addition, Fig.~\ref{fig:supp_industryasm_example} presents more examples to better illustrate the diverse camera viewpoints, actors, backgrounds, and activities in our IndustryASM dataset that are common in manufacturing settings.

\begin{figure}[t]
	\centering
		\includegraphics[width=\linewidth, trim = 0mm 0mm 0mm 0mm, clip]{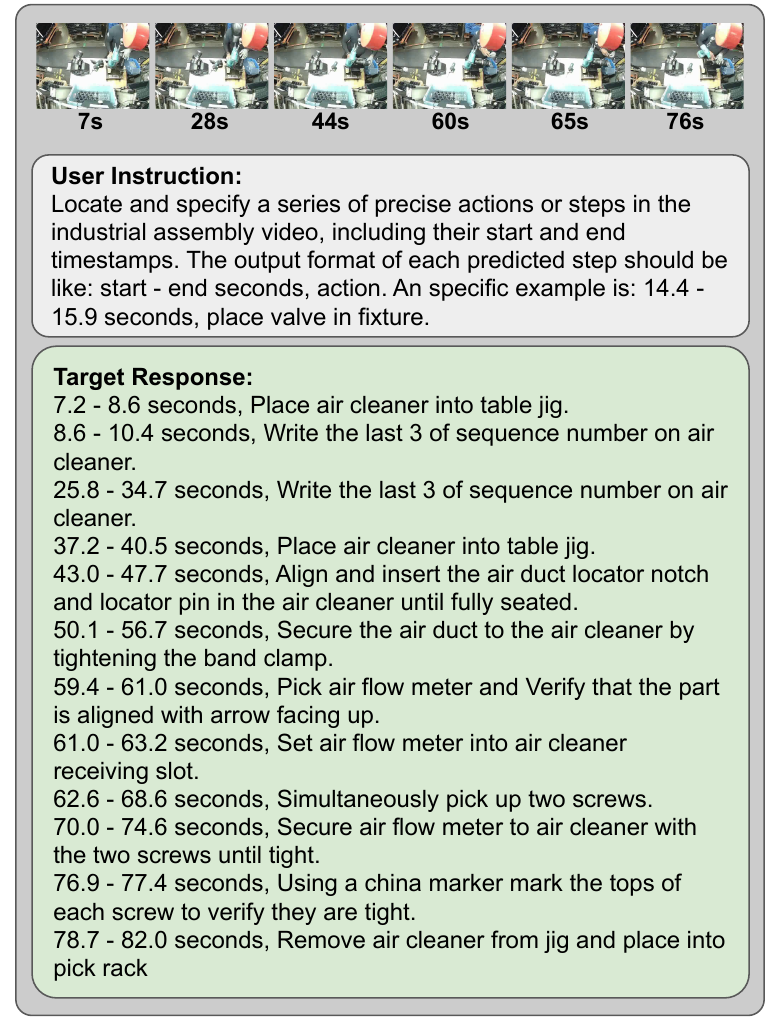}
	\caption{Examples of our instruction and target response for the temporal action segmentation task on an IndustryASM video.}
	\label{fig:supp_industryasm}
\end{figure}

\subsection{Annotation Process}
Our IndustryASM videos are annotated with action names and action timestamps by industrial engineers. To ensure quality, two labelers are assigned for each video, i.e., one labeler provides the labels for the video, while another labeler checks the labels. If there are conflicts, both labelers discuss to fix them. In general, roughly 8\% of the videos have conflicts and need fixing, resulting in an agreement rate of around 92\% between labelers. We manually write the user instructions for the temporal action segmentation task and generate the ground truth responses by using the action names and action timestamps. Fig.~\ref{fig:supp_industryasm} illustrates examples of our instruction and ground truth response for temporal action segmentation on an IndustryASM video.

\subsection{Maintenance and Release Plans}
We host our IndustryASM dataset on Google Drive. We will release version updates to incorporate any corrections and add additional data/annotations if and when available. These updates will follow a versioning scheme (e.g., v1.0) and there will be documentations on any changes in the release notes. Our dataset can be downloaded at \url{https://retrocausal.ai/research/} and we plan to actively maintain it for at least 5 years following the release date.

\section{Implementation Details}
\label{sec:supp_implementation}

We perform grid search to find the best hyperparameter settings for TemporalVLM on supervised dense video captioning on YouCook2~\cite{zhou2018towards} and use them for all the remaining experiments. We present our hyperparameter settings in Tab.~\ref{tab:supp_hyperparams}. We sample $C = 6$ clips from the video, with each consisting of $N^c_f = 96$ frames. The frame encoder processes frames of sizes $224 \times 224$ across $14 \times 14$ patches. Instead of sending timestamps from across the entire video, we send clip-wise timestamps along with their corresponding frames to the frame encoder and clip encoder respectively. Inside the clip encoder, we divide the timestamp encoded input to the video Q-former into windows of size $q = 32$ frames and overlap $o = 16$ frames. We then fuse the concatenated window-wise outputs of the Q-former via multi-headed self-attention with 8 attention heads that have the same hidden size $N_v$ as the Q-Former.  The output of the clip encoder contains a fused time-aware representation of each clip. 

We then concatenate the time-aware clip features according to their temporal order of appearance in the video. This sequence is of dimensions $(C \times N^c_f,~N_{V})$, with $N_V$ denoting the dimension of the video tokens. This sequence is passed to the BiLSTM module which uses information from both past and future states to output a global representation of dimensions $(C \times N^c_f,~2 \times N_{V})$. The BiLSTM module comprises of two hidden layers, with one hidden layer for the forward LSTM and another hidden layer for the backward LSTM. Projection layers are used to project the BiLSTM output into the LLM input space. The first layer projects the BiLSTM output to the dimensions of $(C \times N^c_f,N_{LLM})$ and then the second layer projects the output of the first layer to the sizes of $(N^c_f,N_{LLM})$ which are the input dimensions required by the LLM.

\begin{table}[!t]
  \footnotesize
  \centering
  \begin{tabular}{|l|l|}
    \hline
    \textbf{\footnotesize{Hyperparameter}} & \textbf{\footnotesize{Value}} \\ \hline
    Number of clips $C$ & 6 \\ \hline
    Number of frames sampled per clip $N^c_f$ & 96 \\ \hline
    Frame encoder patch size & 14$\times$14 \\ \hline
    Frame resolution & 224$\times$224  \\ \hline
    Frame sampling type & uniform \\ \hline

    Number of epochs & 7 \\ \hline
    Batch size & 32 \\ \hline
    Learning rate & 5e-5 \\ \hline
    Warmup learning rate & 5e-6 \\ \hline
    Weight decay & 0.01 \\ \hline
    Optimizer & Adam \\ \hline
    AdamW $\beta$ & (0.9,0.999) \\ \hline
    
    Q-Former window size $q$ & 32 \\ \hline
    Overlap $o$ & 16 \\ \hline
    Number of visual tokens per window & 32 \\ \hline
    Number of layers in clip encoder &2  \\ \hline
    Number of layers in image encoder & 12 \\ \hline
    Number of attention heads in fusion module & 8 \\ \hline
    Q-Former hidden size $N_{V}$ & 768 \\ \hline
    Fusion module hidden size $D$ & 768 \\ \hline
    BiLSTM input size & 768 \\ \hline
    BiLSTM hidden size & 768 \\ \hline
    LLM hidden size $N_{LLM}$ & 4096 \\ \hline

  \end{tabular}
  \caption{Hyperparameter settings.}
  \label{tab:supp_hyperparams}
\end{table}

\section{Effects of Number of Training Epochs}
\label{sec:supp_no_of_epochs}
Tab.~\ref{table:supp_three_epoch} shows the performance of our model across different training epochs. Since our model uses a learnable fusion module in the clip encoder and a learnable BiLSTM module to aggregate clip-level features into a global representation, we observe the best results by training for 7 epochs. Tab.~\ref{table:supp_three_epoch} reports our results with 3 epochs, which are worse than ours with 7 epochs but still outperform those of TimeChat~\cite{ren2024timechat} and LongVLM~\cite{weng2024longvlm}.

\begin{table*}[t]
\small
\centering
\renewcommand{\arraystretch}{1.0}
\begin{tabular}{@{}l@{\hspace{9pt}}clll|ll|ll@{}}
\toprule
\multirow{3}{*}{\textbf{Model}} 
& \multirow{3}{*}{\textbf{Epochs}} 
& \multicolumn{3}{c|}{\textbf{Dense Captioning}} 
& \multicolumn{2}{c|}{\textbf{Highlight Detection}} 
& \multicolumn{2}{c}{\textbf{Temporal Grounding}} \\ \\
                 &  & \multicolumn{3}{c|}{\textbf{YouCook2}}                 & \multicolumn{2}{c|}{\textbf{QVHighlights}}               & \multicolumn{2}{c}{\textbf{Charades-STA}}              \\ \cmidrule(l){3-9} 
                 &  & \textbf{SODA\_c}   & \textbf{CIDEr}        & \textbf{F1}          & \textbf{mAP}                  & \textbf{HIT@1}        & 
                 \textbf{R@1 \textsubscript{(IoU=0.5)}}             
                 & 
                 \textbf{R@1 \textsubscript{(IoU=0.7)}}   
                 \\ \midrule
                 
TimeChat \textsubscript{\cite{ren2024timechat}}  & 3 & 3.1 & 10.3 & 19.5 & 21.7 & 37.9 & 46.7 & 23.7 \\
LongVLM \textsubscript{\cite{weng2024longvlm}}  & 3 & 2.3 & 8.1 & 16.9 & 16.0 & 22.5 & 27.2 & 11.9 \\
TemporalVLM  & 3 & \underline{\textit{3.2}} & \underline{\textit{12.9}} & \underline{\textit{19.7}} & \underline{\textit{23.9}} & \underline{\textit{42.3}} & \underline{\textit{50.0}} & \underline{\textit{25.9}} \\
TemporalVLM  & 7& \textbf{3.4} & \textbf{13.2} & \textbf{20.0} & \textbf{25.1} & \textbf{43.0} & \textbf{54.4} & \textbf{29.0} \\

\bottomrule
\end{tabular}%
\caption{Effects of number of training epochs. Best results are in \textbf{bold}, while second best ones are \underline{\textit{underlined}}.}
\label{table:supp_three_epoch}
\end{table*}

\section{General Video Understanding Results}
\label{sec:supp_benchmark}

Following LongVLM~\cite{weng2024longvlm}, we evaluate the performance of our TemporalVLM model on the general video understanding benchmark provided by Video-ChatGPT~\cite{maaz2023video} in Tab.~\ref{tab:supp_benchmark}. The evaluation metrics include Correctness Information (CI), Detail Orientation (DO), Contextual Understanding (CU), Temporal Understanding (TU), and Consistency (C). It is evident from Tab.~\ref{tab:supp_benchmark} that our TemporalVLM model achieves the best performance across all metrics, outperforming all competing methods, including the original LongVLM model.

\begin{table*}[!t]
\small
\begin{center}

\begin{tabular}{l|c|ccccc}
\toprule
\textbf{Model} & \textbf{Data} & \textbf{CI}   & \textbf{DO}   & \textbf{CU}   & \textbf{TU}   & \textbf{C}  \\ \midrule
Video-LLaMA\textsubscript{~\cite{zhang2023video}}      & 10M                  & 1.96          & 2.18          & 2.16          & 1.82          & 1.79          \\
Video-ChatGPT\textsubscript{~\cite{ren2024timechat}}    & 100K                 & 2.50          & 2.57          & 2.69          & 2.16          & 2.20          \\
Valley\textsubscript{~\cite{luo2023valley}}  & 234K & 2.43 & 2.13 & 2.86 & 2.04 & 2.45 \\
BT-Adapter\textsubscript{~\cite{liu2024bt}}       & 10M             & 2.68 & 2.69   & 3.27 & 2.34    & 2.46   \\
LongVLM\textsubscript{~\cite{weng2024longvlm}}  & 100K                 & \underline{\textit{2.76}}    & \underline{\textit{2.86}} & \underline{\textit{3.34}} &  \underline{\textit{2.39}} & \underline{\textit{3.11}} \\ 
TemporalVLM   &  100K  & \textbf{2.88}  & \textbf{2.91}  & \textbf{3.45}  & \textbf{2.50}   & \textbf{3.16} \\ \bottomrule
\end{tabular}

\caption{General video understanding results. Best results are in \textbf{bold}, while second best ones are \underline{\textit{underlined}}.}
\label{tab:supp_benchmark}
\end{center}
\end{table*}

\section{Model Size Comparisons}
\label{sec:supp_model_size}

Tab.~\ref{tab:supp_Trainable Parameters} compares the sizes of our TemporalVLM model, TimeChat~\cite{ren2024timechat}, and LongVLM~\cite{weng2024longvlm} in terms of the number of learnable parameters (measured in millions). Our TemporalVLM model includes a learnable fusion module and a learnable BiLSTM module, leading to a 5\% and 4\% increase in the number of trainable parameters over TimeChat and LongVLM respectively. Nevertheless, our TemporalVLM model achieves the best performance across various temporal reasoning and fine-grained understanding tasks, despite using less training data than TimeChat. In particular, TimeChat is trained on the complete TimeIT~\cite{ren2024timechat} and Valley~\cite{luo2023valley} datasets, whereas our TemporalVLM model is trained on a subset\footnote{We could not download YT-Temporal~\cite{zellers2022merlot} due to its large size and the restricted number of downloads.} of the TimeIT and Valley datasets.

\begin{table}[!t]
\small
  \centering
  \begin{tabular}{l c}
    \toprule
    \textbf{\footnotesize{Model}} & \textbf{\footnotesize{Parameters}} \\ \midrule
    TimeChat \textsubscript{\cite{ren2024timechat}} &  241M \\ 
    LongVLM \textsubscript{\cite{weng2024longvlm}} &  244M \\ 
    TemporalVLM  &   255M \\  
    \bottomrule
  \end{tabular}
  \caption{Model size comparisons.}
  \label{tab:supp_Trainable Parameters}
\end{table}


\section{Run Time Comparisons}
\label{sec:supp_inference_time}

We compare the inference times (measured in seconds) of different models on three tasks, i.e., Dense Video Captioning (DVC) on YouCook2~\cite{zhou2018towards}, Video Highlight Detection (VHD) on QVHighlights~\cite{lei2021detecting}, and Temporal Video Grounding (TVG) on Charades~\cite{gao2017tall} in Tab.~\ref{table:supp_inference_time}. From the results, TimeChat~\cite{ren2024timechat} is about 35\% more efficient than LongVLM~\cite{weng2024longvlm} and TemporalVLM, since both LongVLM and TemporalVLM perform clip sampling and clip encoding.

\begin{table}[t!]
\centering
\small
\begin{tabular}{@{}llll@{}}
\toprule
\textbf{Model} & \textbf{DVC} & \textbf{VHD} & \textbf{TVG} \\ \midrule
TimeChat \textsubscript{\cite{ren2024timechat}}    & 10.1s & 5.6s & 3.7s \\
LongVLM \textsubscript{\cite{weng2024longvlm}}     & 13.2s & 7.3s & 4.9s \\ 
TemporalVLM & 13.5s & 7.6s & 5.1s \\
\bottomrule
\end{tabular}
\caption{Run time comparisons.}
\label{table:supp_inference_time}
\end{table}

\section{FLOPS Comparisons}
\label{sec:flops_comparison}
We further compute the number of floating point operations per second (FLOPS) during the forward pass of the model as a measure of runtime efficiency. As observed from Tab.~\ref{table:supp_flops}, TemporalVLM conducts 4.46\% more flops than LongVLM~\cite{weng2024longvlm} and 16.83\% more flops than TimeChat~\cite{ren2024timechat}, since it relies on clip-level attention based fusion and BiLSTM aggregation.

\begin{table}[!t]
\small
  \centering
  \begin{tabular}{l c}
    \toprule
    \textbf{\footnotesize{Model}} & \textbf{\footnotesize{FLOPS/iteration}} \\ \midrule
    TimeChat \textsubscript{\cite{ren2024timechat}} &  $1.01 \times 10^{13}$\\ 
    LongVLM \textsubscript{\cite{weng2024longvlm}} &  $1.13 \times 10^{13}$ \\ 
    TemporalVLM  &   $1.18 \times 10^{13}$ \\  
    \bottomrule
  \end{tabular}
  \caption{FLOPS/iteration comparisons.}
  \label{table:supp_flops}
\end{table}

\section{Effects of Number of Short-Term Clips}
\label{sec:supp_no_of_clips}
Tab.~\ref{table:ablation_clips} presents the effects of number of short-term clips on our TemporalVLM model. Due to memory limitations, we have experimented with three values for the number of short-term clips, namely 2, 4, and 6. It is evident from Tab.~\ref{table:ablation_clips} that the performance is improved with increasing the number of short-term clips, since our model is able to access more data from the input video.

\begin{table}[t!]

\centering
\small
\begin{tabular}{@{}llll@{}}
\toprule
\textbf{Clips} &  \textbf{SODA\_c}   & \textbf{CIDEr}  & \textbf{F1} \\ \midrule
2 & 2.6 & 9.2& 18.5 \\
4 & \underline{\textit{3.1}} & \underline{\textit{11.3}} & \underline{\textit{18.9}} \\
6 & \textbf{3.4} & \textbf{13.2} & \textbf{20.0} \\ 
\bottomrule
\end{tabular}
\caption{Effects of number of short-term clips. Best results are in \textbf{bold}, while second best ones are \underline{\textit{underlined}}.}

\label{table:ablation_clips}
\end{table}

\section{Results with a More Recent LLM}
\label{sec:supp_llm_models}

\begin{figure}[t]
	\centering
		\includegraphics[width=\linewidth, trim = 0mm 0mm 0mm 0mm, clip]{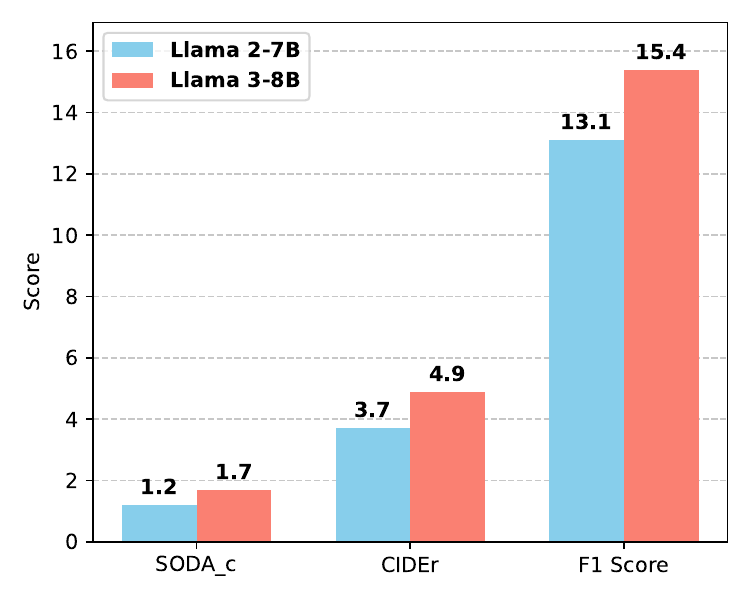}
	\caption{Results with a more recent LLM.}
	\label{fig:supp_llama_model_comparison}
\end{figure}

The performance gain achieved by employing a more advanced LLM as the language decoder in our TemporalVLM model is illustrated in Fig.~\ref{fig:supp_llama_model_comparison}. As LLama 3-8B is pretrained on a significantly larger and diverse dataset, it outperforms Llama 2-7B by a significant margin. The results demonstrate that as LLMs improve, the performance of VLMs, including our TemporalVLM model, will also improve alongside it.

\section{Qualitative Results}
\label{sec:supp_qualitative}

\begin{figure*}[t]
	\centering
		\includegraphics[width=1.0\linewidth, trim = 0mm 0mm 0mm 0mm, clip]{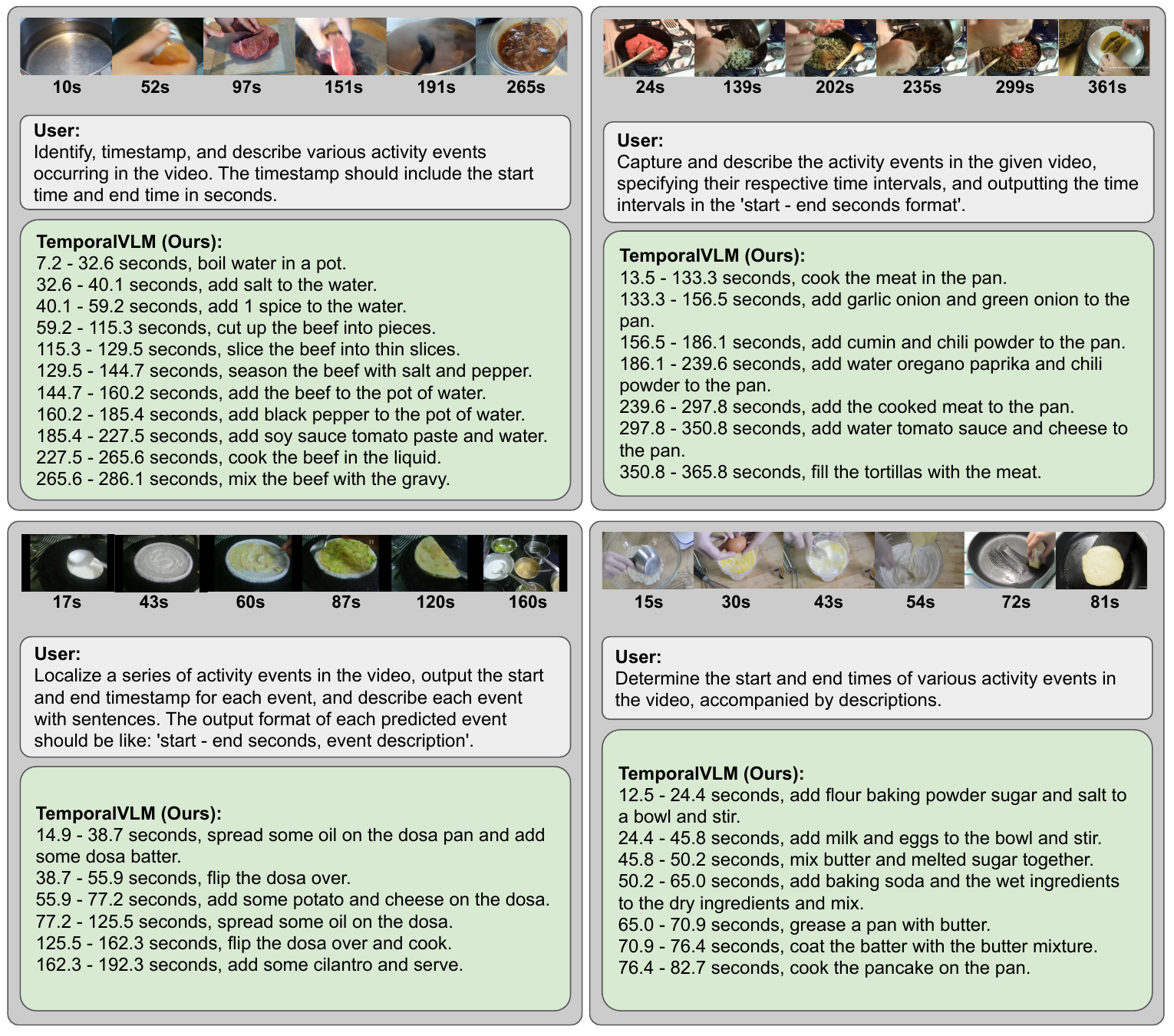}
	\caption{Qualitative examples highlighting the dense video captioning capabilities of TemporalVLM in the supervised setting. The model is asked to provide the timestamps of actions that occur in the videos along with brief descriptions of the actions.}
	\label{fig:supp_dvc}
\end{figure*}

\begin{figure*}[t]
	\centering
		\includegraphics[width=1.0\linewidth, trim = 0mm 0mm 0mm 0mm, clip]{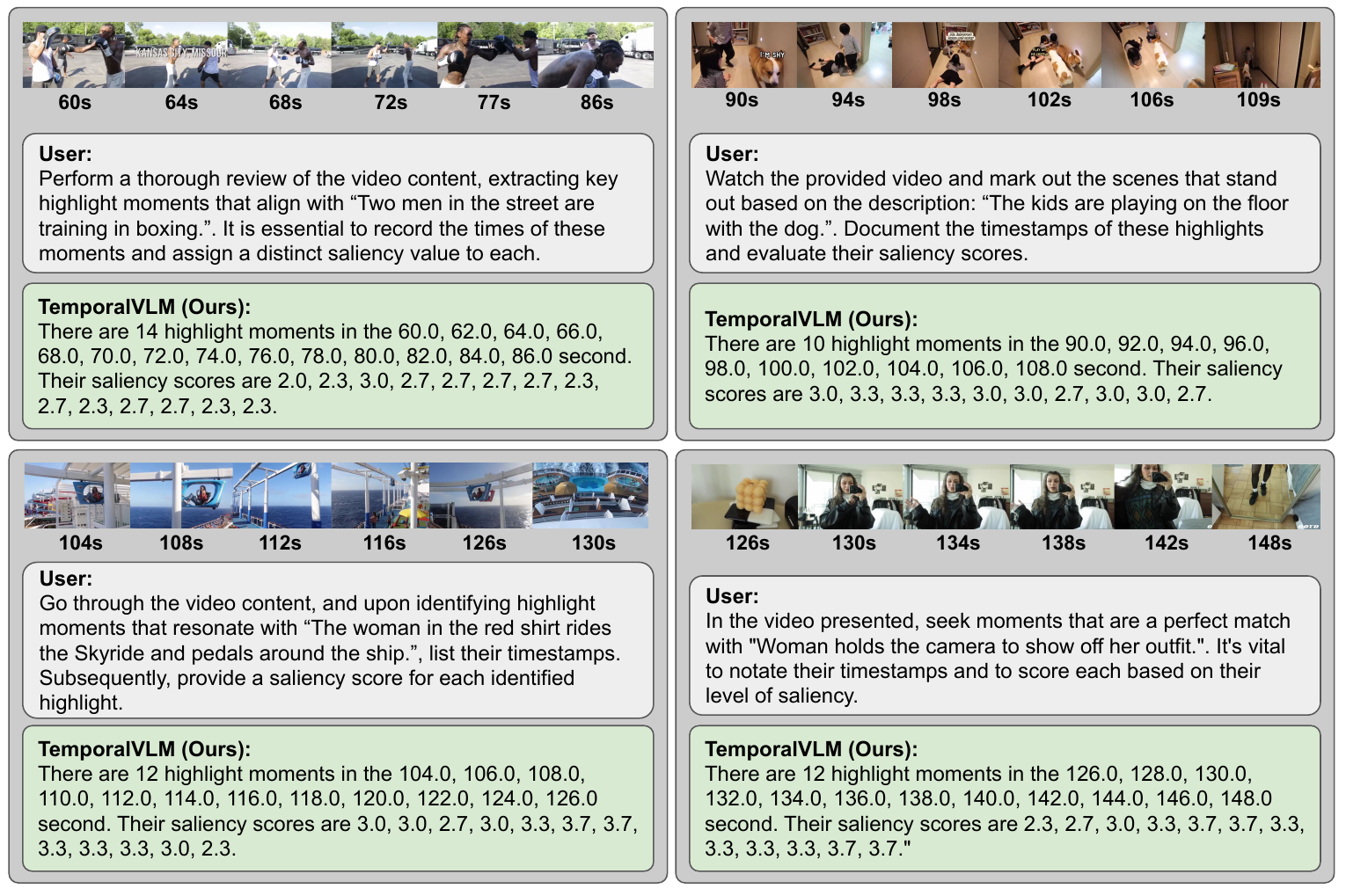}
	\caption{Qualitative examples demonstrating the video highlight detection capabilities of TemporalVLM in the supervised setting. The model is given a video along with an action description. It is prompted to provide the frames that match the action description and their saliency scores.}
	\label{fig:supp_vhd}
\end{figure*}

\begin{figure*}[t]
	\centering
		\includegraphics[width=1.0\linewidth, trim = 0mm 0mm 0mm 0mm, clip]{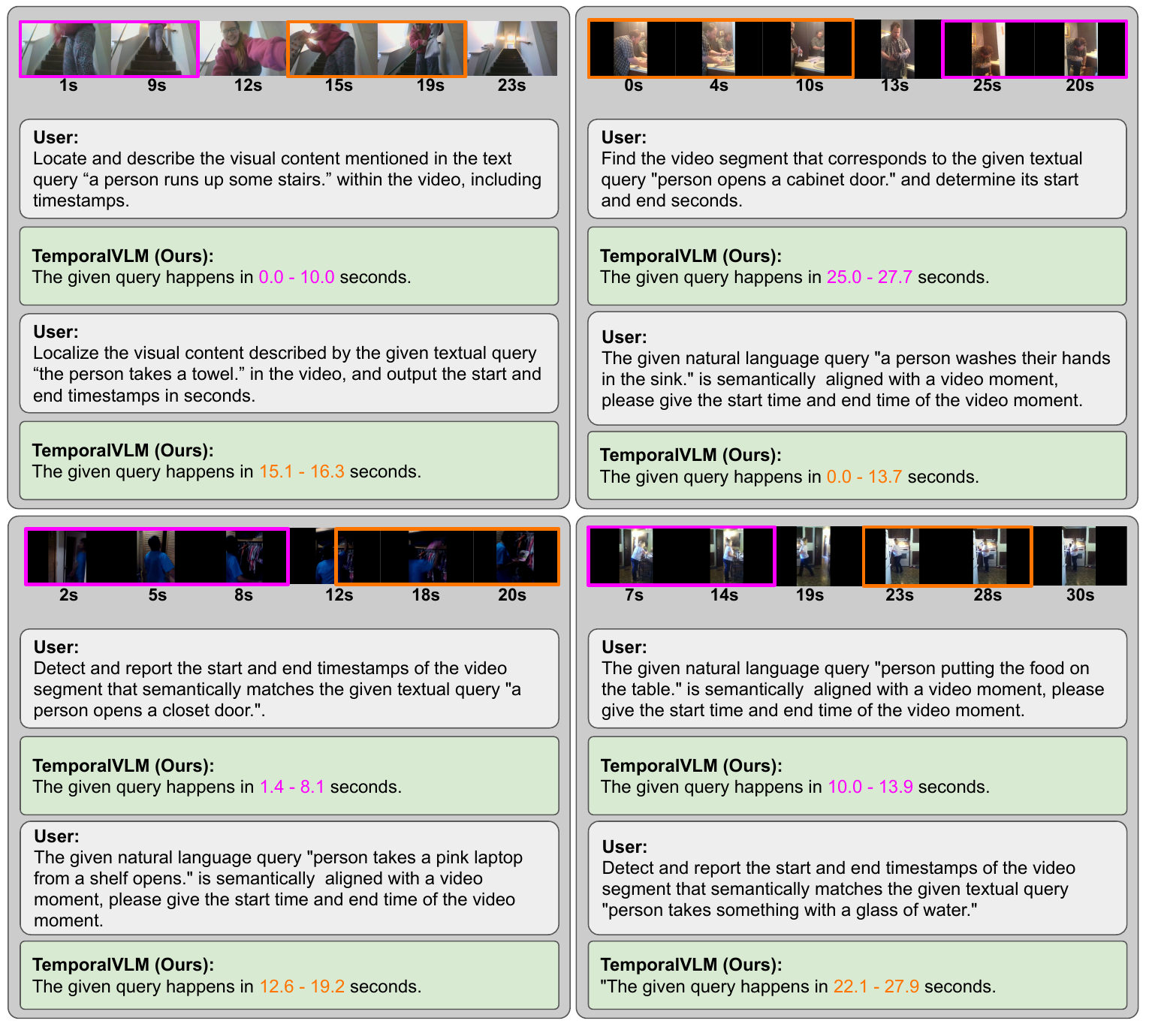}
	\caption{Qualitative examples showing the temporal video grounding capabilities of TemporalVLM in the supervised setting. A video and a query is given to the model. It is prompted to provide the timestamps at which the query occurs. \textbf{\color{magenta} Magenta} represents the ground truth and predicted timestamps for the first query, while \textbf{\color{orange} Orange} indicates those for the second query.}
	\label{fig:supp_tvg}
\end{figure*}

\begin{figure}[t]
	\centering
		\includegraphics[width=1.0\linewidth, trim = 0mm 0mm 0mm 0mm, clip]{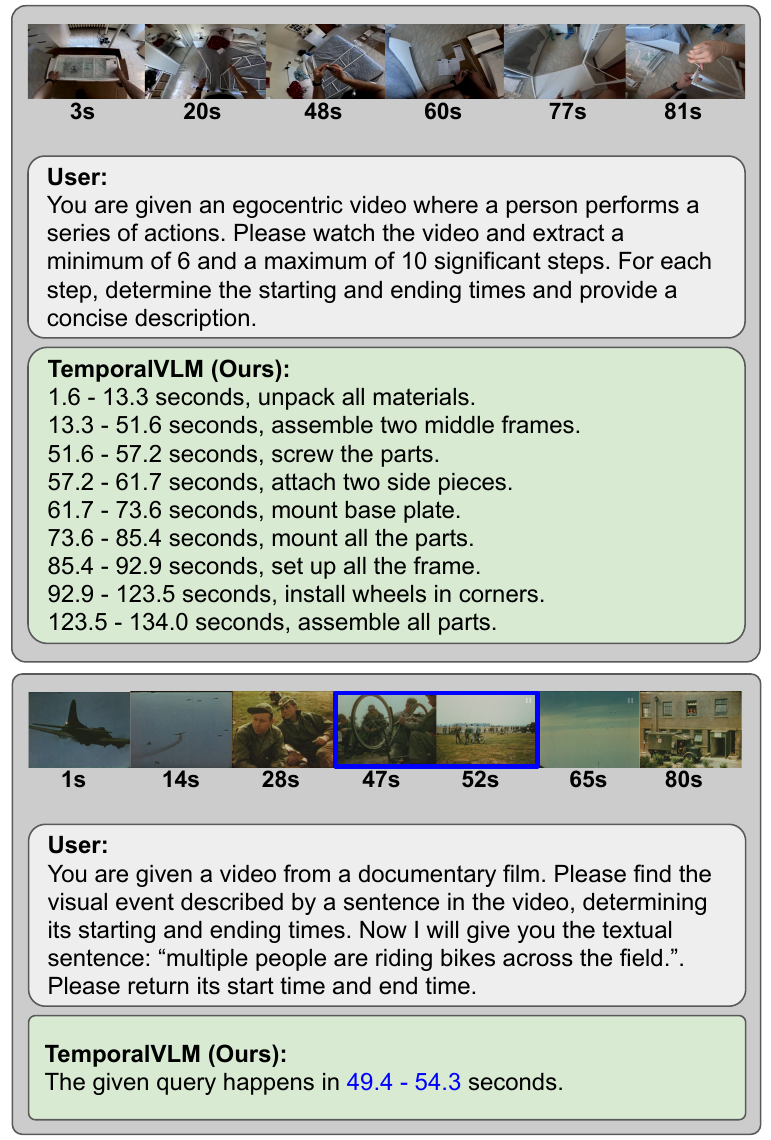}
	\caption{Examples of the generalization capabilities of TemporalVLM in the zero-shot setting. (Top) The model is prompted to provide the timestamps of actions and brief descriptions of actions for an egocentric video of furniture assembly. (Bottom) The model is provided with a documentary film. It is asked to predict the timestamps when the query happens. \textbf{\color{blue} Blue} denotes the ground truth and predicted timestamps.}
	\label{fig:supp_generalization}
\end{figure}

In addition to Figs.~4 and 5 in the main paper, we provide additional qualitative results in this section. In particular, Figs.~\ref{fig:supp_dvc}, \ref{fig:supp_vhd}, and \ref{fig:supp_tvg} present qualitative results by TemporalVLM in the supervised setting for dense video captioning, video highlight detection, and temporal video grounding respectively. More importantly, qualitative results demonstrating the generalization abilities of TemporalVLM in the zero-shot setting are shown in Fig.~\ref{fig:supp_generalization}. Overall, our TemporalVLM model demonstrates promising performance in a variety of temporal reasoning and fine-grained understanding tasks.

\section{Limitations} 
\label{sec:supp_limit}
Despite promising performance in several temporal reasoning tasks, including dense video captioning, temporal video grounding, video highlight detection, and temporal action segmentation, our time-aware video LLM may struggle with complex temporal reasoning tasks and videos with extreme durations. In addition, our TemporalVLM model often has difficulty dealing with small objects, since it does not include an explicit object detector. Our future works will enhance TemporalVLM for tackling complex temporal reasoning tasks, videos with extreme durations, and dealing with small objects.

\section{Societal Impacts} 
\label{sec:supp_societal}
Our time-sensitive video LLM could enable a variety of applications including optimizing and tracking industry assembly processes. More specifically, for assembly process optimization, industrial engineers could utilize TemporalVLM to automatically decompose a video recording of an assembly process into segments as well as generate a brief description for each segment. In addition, TemporalVLM could improve the performance of frontline workers by tracking their assembly process and notify them as soon as they miss a step. Nevertheless, we acknowledge that our time-aware video LLM could be misused for surveillance and monitoring of individuals, which emphasizes the importance of responsible AI principles to guide the use of this technology.

\end{document}